%% file: acl2021.tex
\documentclass[11pt,a4paper]{article}
\usepackage[hyperref]{acl2021}
\usepackage{times}
\usepackage{latexsym}

\usepackage{microtype}

\aclfinalcopy % Uncomment this line for the final submission
\usepackage{graphicx}
\usepackage{bm}
\usepackage{amssymb}
\usepackage{amsmath}
\usepackage[T1]{fontenc}
\usepackage{booktabs}
\usepackage{xcolor}
\usepackage{multirow}
\usepackage{booktabs}
\usepackage{verbatim}
\usepackage{mdwlist}
\usepackage[ruled,norelsize,linesnumbered]{algorithm2e}
\usepackage{amsthm}
\usepackage[capitalize]{cleveref}
\usepackage{paralist}
\usepackage{enumitem}
\usepackage{subcaption}
\usepackage{paralist}

\input{math}
\input{packages}
\usepackage{xspace}
\newcommand{\nmt}{\textsc{nmt}\xspace}
\newcommand{\mt}{\textsc{mt}\xspace}

\newcommand{\ar}{\textsc{ar}\xspace}
\newcommand{\nar}{\textsc{nar}\xspace}
\newcommand{\slkd}{\textsc{slkd}\xspace} % replace all use of distillation with \slkd?
\newcommand{\levt}{LevT\xspace}
\newcommand{\maskt}{MaskT\xspace}
\newcommand{\bleu}{\textsc{bleu}\xspace}

\newcommand{\frs}{\textsc{frs}\xspace}
\newcommand{\ece}{\textsc{ece}\xspace}

\usepackage{enumitem}
\setitemize{itemsep=1pt,topsep=1pt,parsep=1pt,partopsep=1pt}

\title{How Does Distilled Data Complexity Impact the Quality \\and Confidence of Non-Autoregressive Machine Translation?}

\author{Weijia Xu\thanks{~~Work done during internship at Microsoft Research Asia.}$~\dagger~~$ 
  Shuming Ma$\ddagger~~$
  Dongdong Zhang$\ddagger~~$
  Marine Carpuat$\dagger~~$ \\
  $\dagger$Department of Computer Science, University of Maryland \\
  $\ddagger$Microsoft Research Asia  \\
  {\tt \{weijia, marine\}@cs.umd.edu, \{shumma, dozhang\}@microsoft.com} \\
  }

\date{}

\begin{document}
\maketitle
\begin{abstract}
\looseness=-1
While non-autoregressive (\nar) models are showing great promise for machine translation~(\mt), their use is limited by their dependence on knowledge distillation from autoregressive models. To address this issue, we seek to understand why distillation is so effective. Prior work suggests that distilled training data is less complex than manual translations. Based on experiments with the Levenshtein Transformer and the Mask-Predict \nar models on the WMT14 German-English task, this paper shows that different types of complexity have different impacts: while reducing lexical diversity and decreasing reordering complexity both help \nar learn better alignment between source and target, and thus improve translation quality, lexical diversity is the main reason why distillation increases model confidence, which affects the calibration of different \nar models differently.
\end{abstract}

\input{newintro}
\input{method}
\input{experiment}

\section{Conclusion}

We investigated the effect of knowledge distillation in \nar models trained on distilled data that differs along two types of complexity \---\ lexical diversity and degree of word reordering. Reducing lexical diversity and decreasing word reordering degree both boost the confidence of source-target attention, suggesting that they help \nar models learn the alignment between source and target. 
Furthermore, distillation increases model confidence by reducing lexical diversity, which improves calibration for \levt but leads to much worse calibration for \maskt.
These findings reveal a connection between distillation and existing techniques to improve \nar via pre-reordering~\citep{RanLLZ2019} or integrating external alignment information in the source-target attention~\citep{LiLHTQWL2019}.\footnote{These techniques still rely on knowledge distillation for training, while this paper contributes a systematic study of factors that impact the effectiveness of distillation.} 

Our findings are based on experiments on the WMT14 English-German corpus, which is widely used in the literature of \nar translation and has interesting typological properties. While we expect these findings to hold for other tasks  that exhibit similar degrees of reordering and lexical diversity, it remains to be seen to what degree they generalize to other language pairs and data settings.  

We hope that this work will inspire future research on understanding of the positive and negative impact of knowledge distillation on \nar models, as well as of the more advanced approaches to improving \nar by integrating lexical choice and word reordering knowledge. In addition, our work also calls for future work on improving the calibration of \nar models.

\section*{Acknowledgments}
\looseness=-1
We thank the anonymous reviewers, Sweta Agrawal, Naomi Feldman, Zara Harmon, Aquia Richburg, Craig Thorburn, the CLIP lab at UMD for helpful comments, and Eleftheria Briakou for her help on result visualization. This research is supported in part by the Office of the Director of National Intelligence (ODNI), Intelligence Advanced Research Projects Activity (IARPA), via contract \#FA8650-17-C-9117. The views and conclusions contained herein are those of the authors and should not be interpreted as necessarily representing the official policies, either expressed or implied, of ODNI, IARPA, or the U.S. Government. The U.S. Government is authorized to reproduce and distribute reprints for governmental purposes notwithstanding any copyright annotation therein.

\bibliography{anthology,acl2021}
\bibliographystyle{acl_natbib}

\input{appendix}

\end{document}

%% file: math.tex
\usepackage{bm}
\usepackage{amssymb}
\usepackage{amsmath}
\usepackage[ruled,norelsize,linesnumbered]{algorithm2e}
\usepackage{amsthm}
\usepackage{dsfont}

\theoremstyle{definition}

\newcommand{\loss}{\mathcal{L}}
\newcommand{\cond}{\,|\,}

\newcommand{\h}[2]{\mathbb{H}_{#1}\left[ #2 \right] }
\newcommand{\ind}[1]{\mathds{1}\left[ #1 \right] }

\newcommand{\kld}[2]{D_{\mathrm{KL}} \left[ \left. \left. #1 \right|\right| #2 \right] }

\makeatletter
\newcommand{\removelatexerror}{\let\@latex@error\@gobble}
\makeatother

%% file: packages.tex
\usepackage{array}% http://ctan.org/pkg/array
\usepackage{colortbl}

\usepackage{url}

\usepackage{pgfplots}
\pgfplotsset{width=10cm,compat=1.9} % This changes the size of each pgfplot figure to 10 cementers, which is huge; you may use different units (pt, mm, in). The compat parameter is for the code to work on the package version 1.9 or latter.

\definecolor{darkblue}{rgb}{0,0,.5}
\definecolor{darkgreen}{rgb}{0,.5,0}
\definecolor{lightgray}{rgb}{.8,.8,.8}
\definecolor{aliceblue}{rgb}{0.75, 0.75, 1.0}
\definecolor{darkseagreen}{rgb}{0.46, 0.74, 0.46}
\definecolor{alizarin}{rgb}{0.82, 0.1, 0.26}
\definecolor{airforceblue}{rgb}{0.36, 0.54, 0.66}
\definecolor{red_graph}{rgb}{0.98, 0.8, 0.8}
\definecolor{blue_graph}{rgb}{0.8, 0.98, 0.8}
\definecolor{crimson}{HTML}{DC143C}
\definecolor{skyblue}{HTML}{87CEEB}
\definecolor{darkred}{HTML}{8B0000}

%% file: newintro.tex
\section{Introduction and Background}
\vspace{-3pt}

When training \nar models for neural machine translation~(\nmt), sequence-level knowledge distillation~\citep{KimR2016} is key to match the translation quality of autoregressive~(\ar) models~\citep{GuBXLS2018,LeeMC2018,GhazvininejadLLZ2019,GuWZ2019}. Knowledge distillation was first proposed to obtain small \textit{student} models that match the quality of a higher-capacity \textit{teacher} models~\citep{LiangDK2008,HintonVD2015}. Sequence-level knowledge distillation~(\slkd)  trains the student model~$p(\boldsymbol{y} \cond \boldsymbol{x})$ to approximate the teacher distribution~$q(\boldsymbol{y} \cond \boldsymbol{x})$ by maximizing the following objective:
$\loss_\text{SEQ-KD} = - \sum_{\boldsymbol{y} \in  \mathcal{Y}} q(\boldsymbol{y} \cond \boldsymbol{x}) \log p(\boldsymbol{y} \cond \boldsymbol{x}) \approx - \sum_{\boldsymbol{y} \in \mathcal{Y}} \ind{\boldsymbol{y} = \hat{\boldsymbol{y}}} \log p(\boldsymbol{y} \cond \boldsymbol{x})$, where~$\mathcal{Y}$ represents the space of all possible target sequences, and~$\hat{\boldsymbol{y}}$ is the output from running beam search with the teacher model~$q$. 

However, we do not yet have a clear picture for how \slkd impacts \nar training. \citet{RenLTZZL2020} show that \slkd reduces the degree of dependency between target tokens. \citet{GuBXLS2018} hypothesize that \slkd reduces the number of modes in the output distribution (alternative translations for a source). This hypothesis was supported by experiments that use multiway parallel data to simulate the modes~\citep{ZhouNG2019}.  \citet{ZhouNG2019} also investigate the impact of data complexity on \nar translation quality \---\ they generate distilled data of varying complexity with \ar models of different capacity and show that higher-capacity \nar models require more complex distilled data to achieve better translation quality. They further show that generating distilled references with mixture of experts~\citep{Shen2019MoE} improves \nar translation quality. However, training samples can be complex in  different ways, and it remains unclear how different types of data complexity alter the internal working of \nar models and their translation quality. We also anticipate that data complexity may impact the uncertainty and calibration of \nar models \---\ an understudied question, unlike for \ar models~\citep{OttAGR2018,WangTSL2020}. 

\looseness=-1
This paper focuses on two types of data complexity \---\ lexical diversity and degree of word reordering. We expose two state-of-the-art \nar models  (Mask-Predict~\citep{GhazvininejadLLZ2019} and Levenshtein Transformer~\citep{GuWZ2019}) to distilled references of varying complexity on the WMT14 German-English task. 
Experiments show that decreasing reordering complexity and reducing lexical diversity via distillation both help \nar models learn better alignment between source and target and thus improve translation quality. Further analysis shows that knowledge distillation lowers model uncertainty by reducing lexical diversity, which affects the calibration of Mask-Predict and Levenshtein Transformer models in opposite directions.

%% file: method.tex
\section{Generating Diverse Distilled References}
\label{sec:method}

We measure \textbf{distilled corpus complexity} with:
\begin{itemize}
    \item \textbf{Word Reordering Degree} computed by the average fuzzy reordering score~(\frs)~\citep{Talbot2011Fuzzy} over all sentence pairs.
    \frs is an \mt evaluation metric introduced to distinguish significant changes in reordering rules of \mt systems on syntactically distant language pairs. A higher \frs indicates that the hypothesis is more monotonically aligned to the source. \citet{ZhouNG2019} show that distilled data has a higher \frs than the real data which may benefit \nar models.
    \item \textbf{Lexical Diversity} which captures the diversity of target word choices given a source word. We compute the lexical diversity~$LD(d)$ of the distilled corpus~$d$ by averaging the entropy of target words~$y$ conditioned on a source word~$x$ \citep{ZhouNG2019}:~$LD(d) = \frac{1}{|\mathcal{V}_x|} \sum_{x \in \mathcal{V}_x}\h{}{y \cond x}$, where~$\mathcal{V}_x$ denotes the source vocabulary.
\end{itemize}

To isolate the impact of complexity factors, we seek to control the  \textbf{faithfulness}~$F(d)$ of the distilled data $d$ to the real parallel data~$r$. We compute it as the KL-divergence of the alignment distribution between the real data~$r$ and the distilled data~$d$ \citep{ZhouNG2019}:
$F(d) = \frac{1}{|\mathcal{V}_x|} \sum_{x \in \mathcal{V}_x}\kld{p_r(y \cond x)}{p_d(y \cond x)}$.

\paragraph{Distilled Sample Generation}
To encourage diversity according to the corpus-level metrics above, we select distilled references for each source from the $k$-best list of \ar hypotheses,\footnote{This is inspired by sequence-level interpolation~\citep{KimR2016}, but they select hypothesis using BLEU while we use more diverse criteria. We use beam search with $k = 32$.} using instantiations of the following score:
\begin{equation*}
    \text{score}(\boldsymbol{\hat{y}} | \boldsymbol{x}, \boldsymbol{y}) = \lambda \, \text{sim}(\boldsymbol{\hat{y}}, \boldsymbol{y}) + (1 - \lambda) \, \text{cxty}(\boldsymbol{\hat{y}}, \boldsymbol{x})
\end{equation*}
where the similarity $\text{sim}(\boldsymbol{\hat{y}}, \boldsymbol{y})$ measures how faithful the hypothesis~$\boldsymbol{\hat{y}}$ is to the original reference~$\boldsymbol{y}$ and the complexity ~$\text{cxty}(\boldsymbol{\hat{y}}, \boldsymbol{x})$ captures the relationship between the target sequence $\boldsymbol{\hat{y}}$ and source sequence~$\boldsymbol{x}$. 
The similarity function is  the smoothed sentence-level \bleu~\citep{ChenC2014} w.r.t the original reference. We use three different complexity functions:
\begin{inparaenum}[1)]
    \item \textbf{\frs},
    \item \textbf{word-alignment score}\footnote{Sum of the log probabilities of each target word conditioned on its aligned source words given by \textit{fast-align}.} that measures complexity on a word level, and
    \item \textbf{\nmt score}\footnote{Log probability of the target sentence conditioned on the source given by an \ar model.} that measures complexity on a sentence level.
\end{inparaenum}

%% file: experiment.tex
\section{Experimental Settings}
\label{sec:exp}

\paragraph{Set-Up} We use En-De and De-En datasets from WMT14~\citep{BojarWMT2014} with the same preprocessing steps as \citet{GuWZ2019}. We evaluate translation quality with case-sensitive tokenized BLEU,\footnote{\url{https://github.com/pytorch/fairseq/blob/master/fairseq/clib/libbleu/libbleu.cpp}} using the Moses tokenizer.

\paragraph{Models}
We use two state-of-the-art \nar models:
\begin{itemize}
    \item \textbf{Mask-Predict~(\maskt)}~\citep{GhazvininejadLLZ2019} uses a masked language model~\citep{Devlin2019BERT} to generate the target sequence by iteratively masking out and regenerating the subset of tokens that the model is least confident about.
    \item \textbf{Levenshtein Transformer~(\levt)}~\citep{GuWZ2019} generates the target sequence through iterative insertion and deletion steps. 
\end{itemize}
All \ar and \nar models adopt the \emph{base} Transformer architecture~\citep{Vaswani2017}. We train all models using a batch size of~$64,800$ tokens for maximum~$300,000$ steps and select the best checkpoint based on validation perplexity~(see Appendix for details). During inference, we set the maximum number of iterations to~$10$. All word alignments in this paper are generated automatically using \textit{fast-align}~\citep{Dyer2013fastalign}.\footnote{This might introduce alignment errors leading to lower absolute \frs scores than with if we had access to gold manual alignments. However, this measurement noise is unlikely to impact our findings because 1) it is likely to be small on distilled data generated by autoregressive \nmt models, which should be easier to align than original translations, and 2) distilled data versions are expected to be impacted uniformly.}

\section{Preliminary: \slkd Helps \nar Learn Word Alignment}
\label{sec:prereordering}

\begin{table}[!t]
\centering
\scalebox{0.9}{
\begin{tabular}{lccc}
\toprule
& Real & Distill & $\Delta$ \\
\midrule
Original & $24.2$ & $26.6$ & $+2.4$ \\
Reordered & $30.0$ & $29.4$ & $-0.6$ \\
\hline
\end{tabular}}
\caption{\bleu scores on the original WMT14 En-De and the synthetic reordered version. For each task, we compare \levt models trained on real vs. distilled data.}
\vspace{-10pt}
\label{tab:reorder}
\end{table}

\begin{table}[ht]
\centering
\scalebox{0.8}{
\begin{tabular}{lcccll}
\toprule
& \multicolumn{3}{c}{Data Property} & \multicolumn{2}{c}{\bleu $\uparrow$} \\
Data Version & \frs & LexDiv & Faith & \maskt & \levt \\
\midrule
Real & $0.46$ & $0.36$ & $0.0$ & $28.0$ & $27.6$ \\
Distilled & $0.55$ & $0.18$ & $7.9$ & $29.6$ & $30.6$ \\
\midrule
\multicolumn{6}{l}{selection via \bleu }\\
$+0.5$ \nmt & $0.55$ & $0.17$ &	$7.6$ & $29.5$ & $30.6$ \\
$+0.5$ w-align & $0.57$ & $0.18$ &	$7.6$ & $29.2$ & $30.1$ \small{\textcolor{crimson}{\boldmath$\downarrow$}} \\
 $+0.5$ \frs & $0.61$ & $0.19$ &	$7.6$ & $28.8$ \small{\textcolor{crimson}{\boldmath$\downarrow$}} & $29.6$ \small{\textcolor{crimson}{\boldmath$\downarrow$}} \\
\midrule
\multicolumn{6}{l}{selection via \bleu }\\
 $+0.2$ \nmt & $0.55$ & $0.17$ &	$7.8$ & $29.2$ & $30.4$ \\
 $+0.2$ w-align & $0.58$ & $0.18$ &	$7.9$ & $28.7$ \small{\textcolor{crimson}{\boldmath$\downarrow$}} & $30.0$ \small{\textcolor{crimson}{\boldmath$\downarrow$}} \\
 $+0.2$ \frs & $0.64$ & $0.19$ &	$7.8$ & $28.5$ \small{\textcolor{crimson}{\boldmath$\downarrow$}} & $29.7$ \small{\textcolor{crimson}{\boldmath$\downarrow$}} \\
\hline
\end{tabular}}
\caption{Translation quality on WMT14 De-En. In the bottom two groups, models are trained on distilled data with similar faithfulness~(\textit{Faith}) but varying degree of reordering~(\textit{\frs}) and lexical diversity~(\textit{LexDiv}). \textcolor{crimson}{\boldmath$\downarrow$} marks significant drops compared to the first row in each group based on the paired bootstrap test at~$p < 0.05$ \citep{Clark2011}.}
\label{tab:quality}
\end{table}

\looseness=-1
Our work is motivated by the hypothesis that \slkd helps \nar models learn (implicit) alignment between source and target words.
We first test this hypothesis by evaluating the effect of \slkd on two datasets:
\begin{inparaenum}[a)]
    \item En-De train/dev/test sets from WMT14, and
    \item a synthetic version of the same task, where word alignment information is embedded by pre-reordering the source words so that they are monotonically aligned with target words (in train/dev/test sets).
\end{inparaenum}

While \slkd improves \bleu by~$+2.4$ on the original En-De task, it has no benefit on the synthetic task (Table~\ref{tab:reorder}).  This supports our hypothesis and is consistent with other findings on real data: \citet{GhazvininejadLLZ2019} and \citet{GuWZ2019} showed that \slkd improves the quality of \nar models more on syntactically distant language pairs such as German-English than on Romanian-English. Furthermore, \citet{RanLLZ2019} showed that automatically pre-reordering the source words improves the translation quality of \nar models. However, unlike in our experiment, \slkd is still needed in real translation scenarios, as exactly pre-ordering the source is not feasible at test time. Thus, we turn to understanding how distilled data helps \nar models on real translation tasks.

\section{Reduced Lexical Diversity in \slkd Improves Translation Quality}
\label{sec:translation_quality}

We have shown that, similar to the effect of pre-reordering, \slkd benefits \nar training by reducing the difficulty of learning the source-target alignment. However, apart from the word reordering degree, reducing the lexical diversity on the target side can also reduce the difficulty of learning the alignment. In this section, we investigate how the two types of data complexity affect how well \nar models capture the source-target alignment, and therefore translation quality.

\slkd impacts both complexity types: the first two rows of Table~\ref{tab:quality} show that \slkd increases \frs by $+0.09$, reduces lexical diversity by $-0.18$, and boosts the \bleu of \maskt and \levt by~$1.6$--$3.0$ over their counterparts trained on real data.

We then compare \nar models trained on distilled data with varying degree of reordering and lexical diversity while controlling for faithfulness (2nd and 3rd group of rows in Table~\ref{tab:quality}). While the absolute BLEU deltas are small, \bleu decreases significantly as the lexical diversity increases despite reduced degree of reordering. This indicates that increased lexical diversity prevails over the effect of lower degree of reordering in decreasing \bleu scores.

\begin{table}[!t]
\centering
\scalebox{0.9}{
\begin{tabular}{lccc}
\toprule
& Acc & Conf & \ece $\downarrow$ \\
\midrule
\ar Transformer & $63.9$ & $72.3$ & $10.34$ \\
\maskt w/o \slkd & $63.7$ & $74.2$ & $10.49$ \\
\maskt w/ \slkd & $65.1$ & $86.5$ & $21.41$ \\
\levt w/o \slkd & $66.8$ & $53.3$ & $20.26$ \\
\levt w/ \slkd & $65.9$ & $71.3$ & $15.17$ \\
\hline
\end{tabular}}
\caption{Average token-level accuracy~(\textit{Acc}), confidence~(\textit{Conf}), and inference \ece~(\textit{\ece}) of \ar and the two \nar models trained with and without \slkd.}
\vspace{-10pt}
\label{tab:infece}
\end{table}

\begin{figure}[!t]
    \centering
    \subfloat[Impact of lexical diversity]{{
        \includegraphics[width=0.38\textwidth]{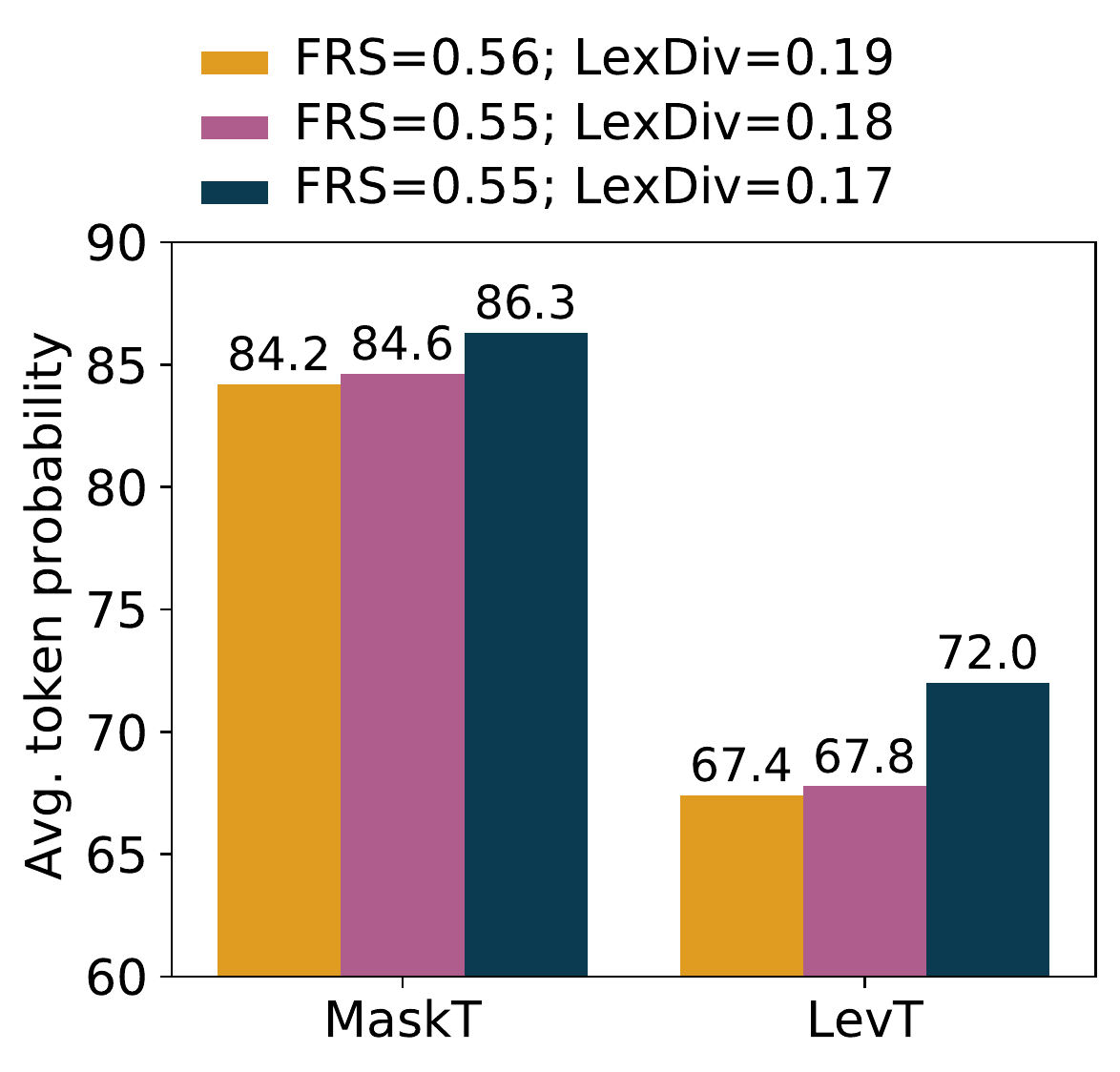}
    }\label{fig:confidence_lex}
    }\hspace{0.1em}
    \subfloat[Impact of word reordering]{{
        \includegraphics[width=0.38\textwidth]{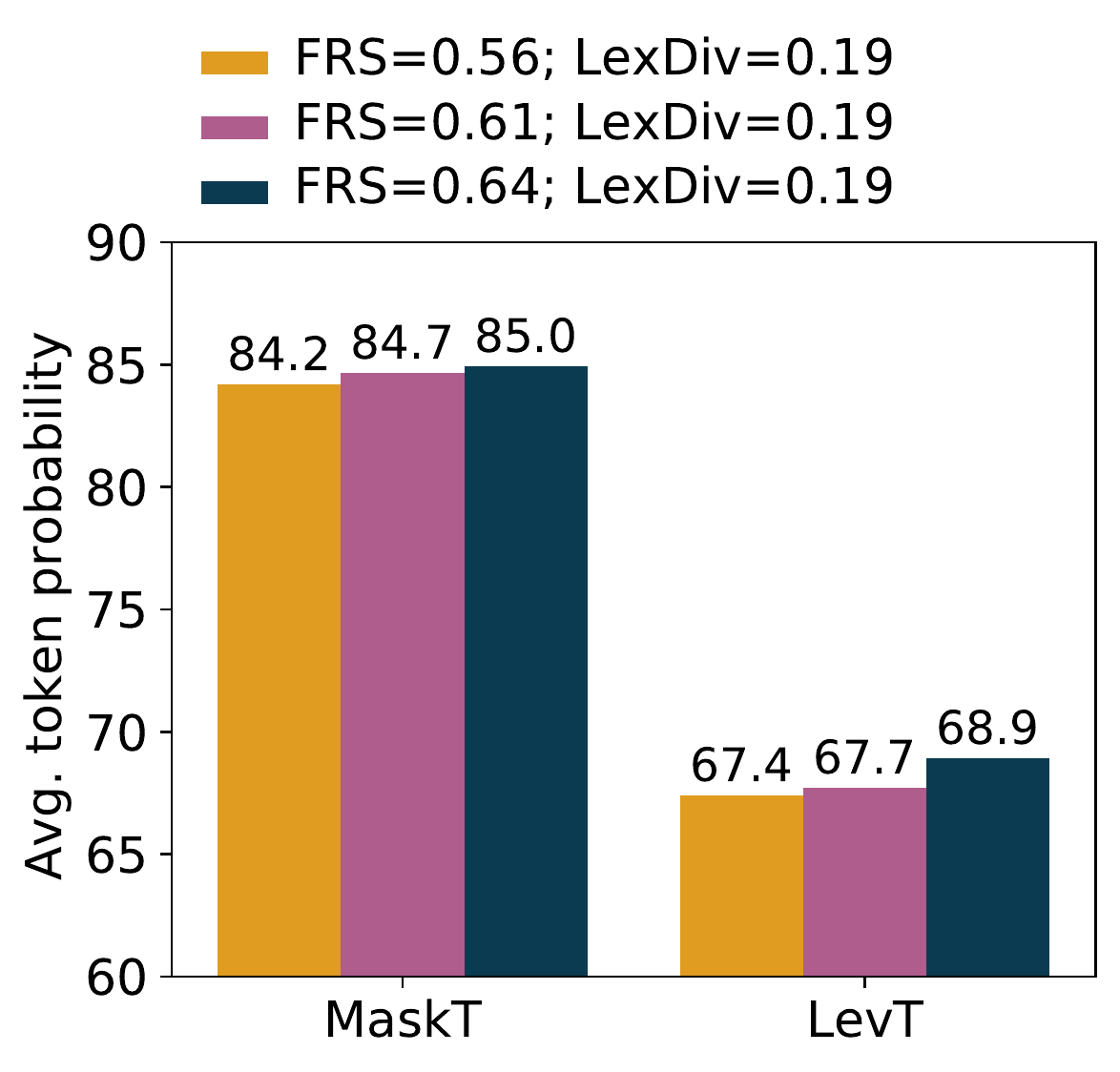}
    }\label{fig:confidence_frs}
    }\\
\caption{Average token-level uncertainty of \maskt and \levt trained on distilled data with decreasing degree of lexical diversity~(a) and word reordering~(b) from yellow to blue.} 
\vspace{-10pt}
\label{fig:tok_confidence_complexity}
\end{figure}

\section{\slkd Increases Confidence of Source-Target Attention}
\label{sec:attention}

To better understand how \slkd helps \nar learn the alignment between source and target, we measure how the \textit{confidence} of the source-target attention changes over decoding iterations.
Following \citet{VoitaTMST2019}, we define the \textit{confidence} of attention heads as the average of the maximum attention weights over source tokens, where the average is taken over target tokens.
Higher confidence scores indicate that the model is more certain about which parts of the source sequence to attend to when predicting the target tokens.

\begin{figure}[!t]
    \centering
    \begin{subfigure}[b]{0.45\textwidth}
        \centering
        \includegraphics[width=\textwidth]{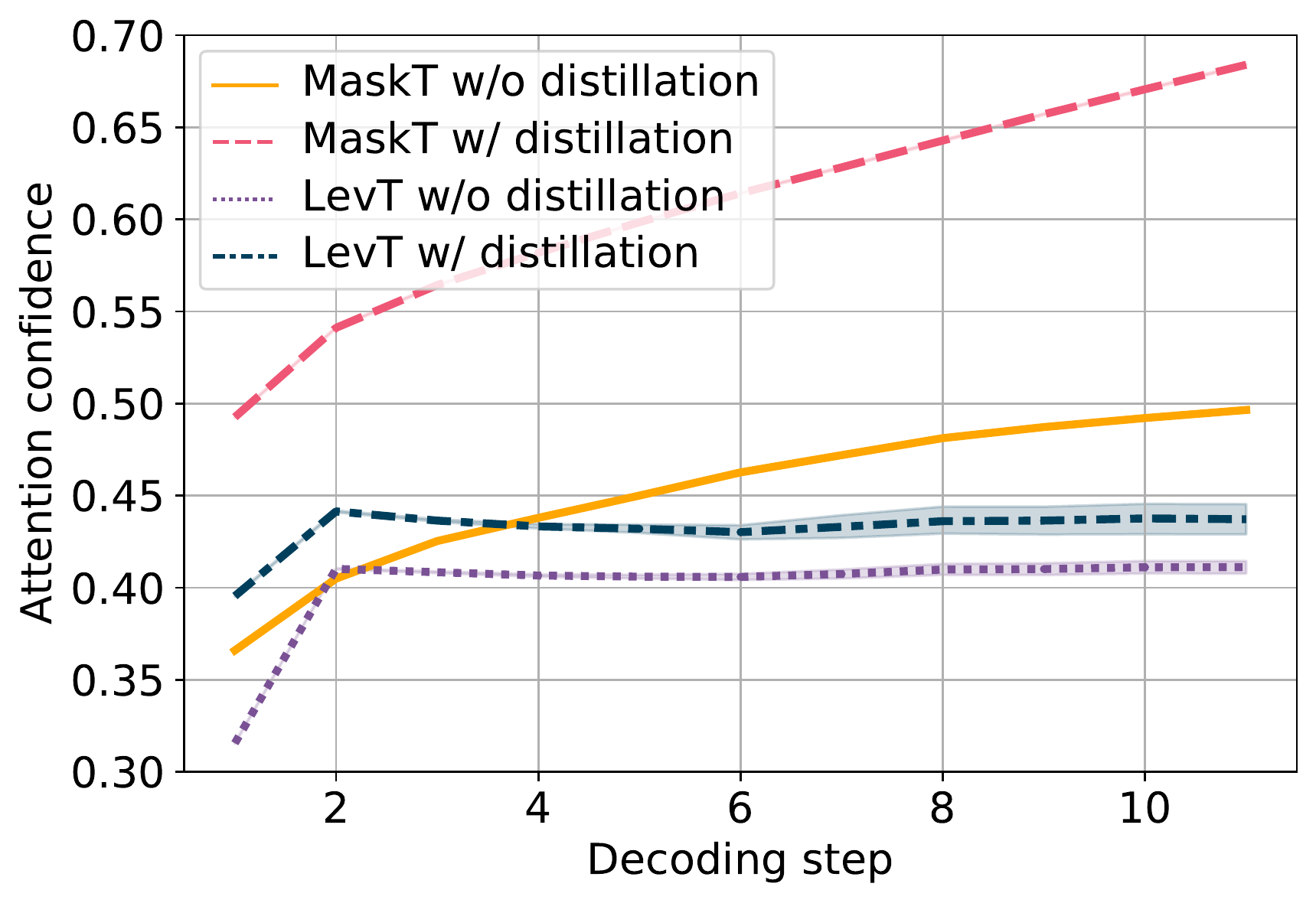}
    \end{subfigure}
\caption{Source-target attention confidence of \levt and \maskt as a function of decoding step.}
\label{fig:attention_nar}
\vspace{-10pt}
\end{figure}

\begin{figure}[!t]
    \centering
    \subfloat[LexDiv on \maskt]{{
        \includegraphics[width=0.4\textwidth]{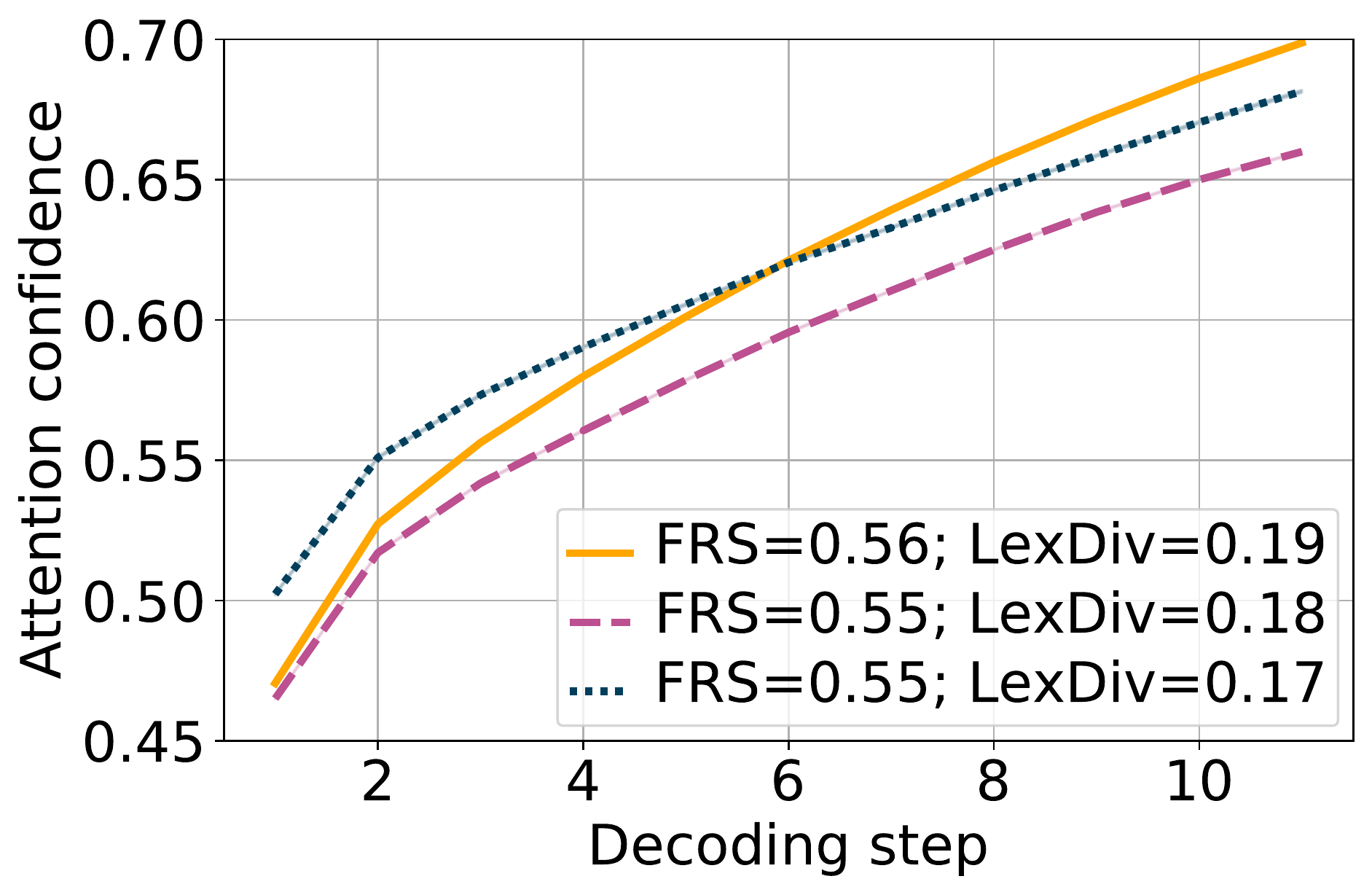}
    }\label{fig:maskT_attention_lex}
    }\hspace{0.1em}
    \subfloat[\frs on \maskt]{{
        \includegraphics[width=0.4\textwidth]{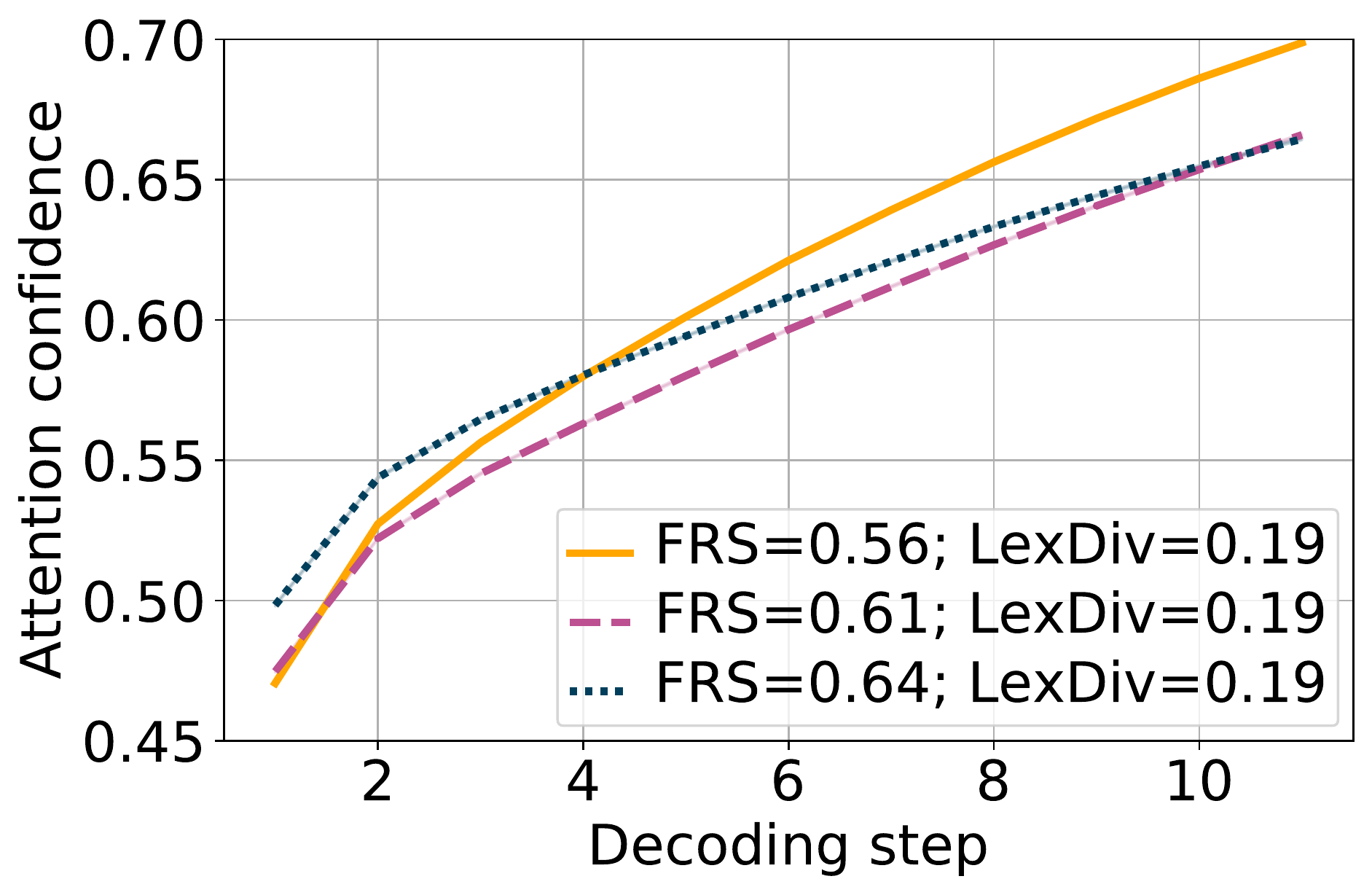}
    }\label{fig:maskT_attention_frs}
    }\\

    \subfloat[LexDiv on  \levt]{{
        \includegraphics[width=0.4\textwidth]{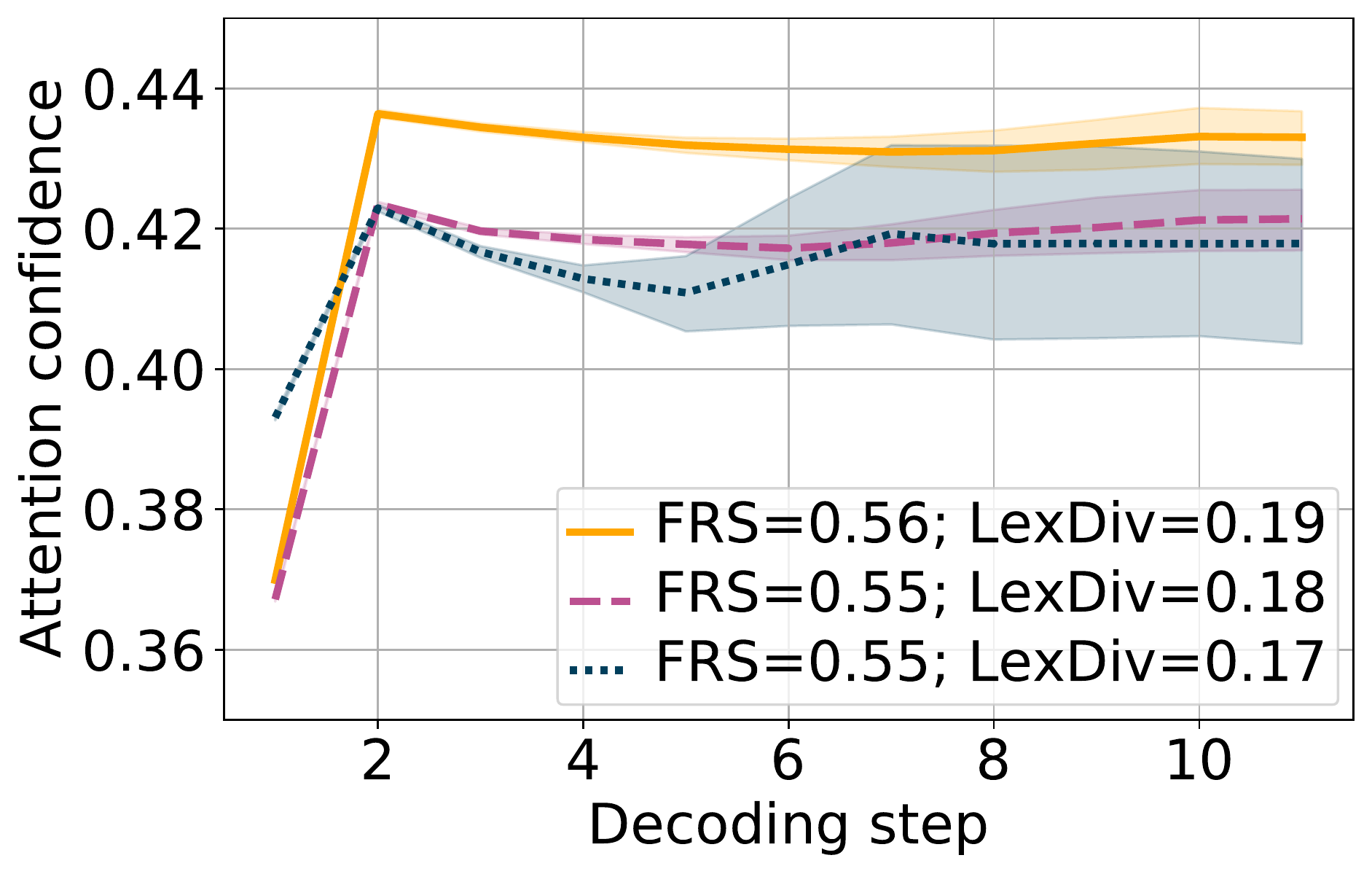}
    }\label{fig:levT_attention_lex}
    }\hspace{0.1em}
    \subfloat[\frs on \levt]{{
        \includegraphics[width=0.4\textwidth]{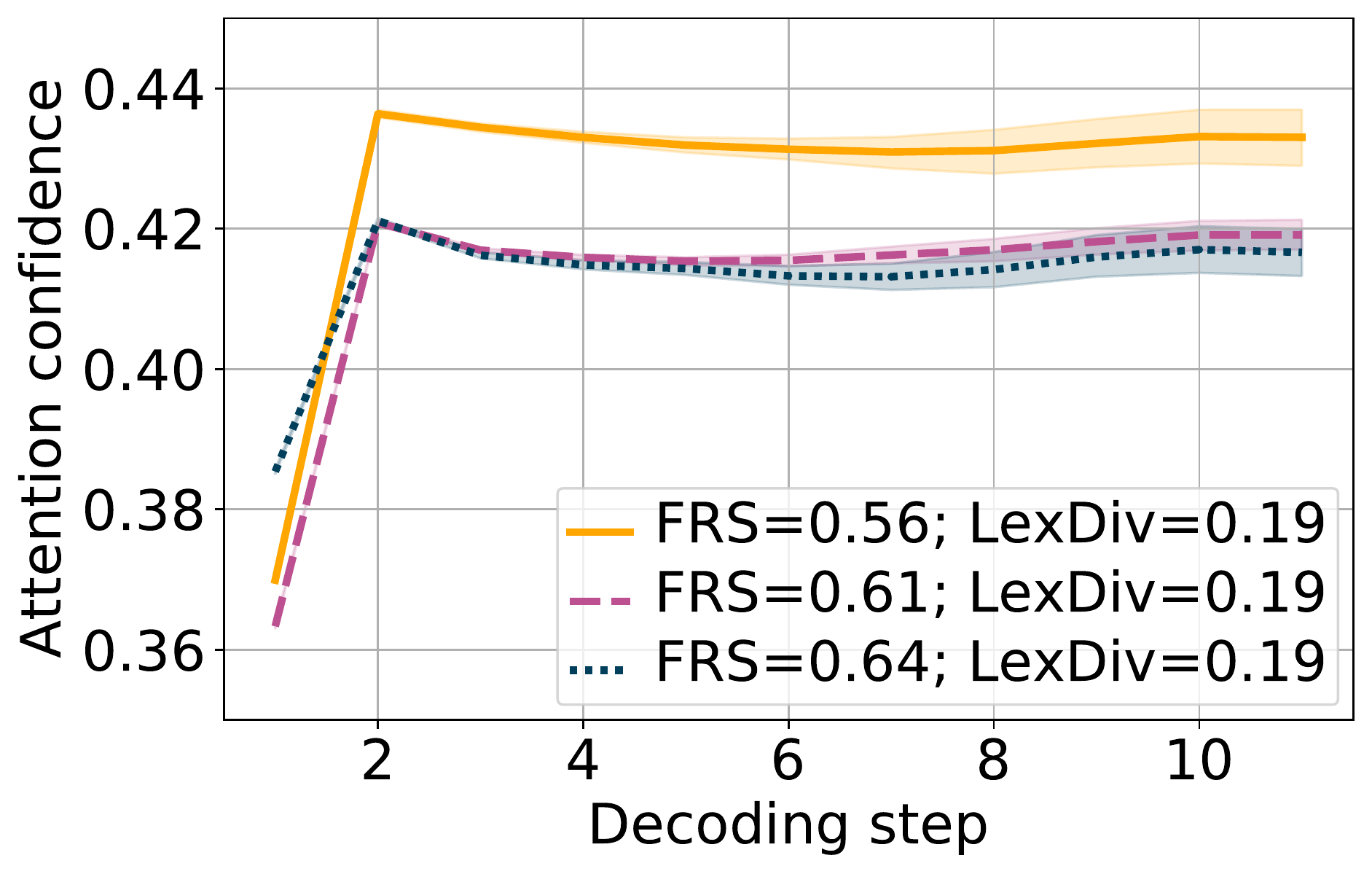}
    }\label{fig:levT_attention_frs}
    }
\caption{Source-target attention confidence as a function of decoding step comparing \maskt and \levt trained on distilled data with varying degree of lexical diversity~(a, c) and word reordering~(b, d).} 
\vspace{-10pt}
\label{fig:attention_complexity}
\end{figure}

As seen in Figure~\ref{fig:attention_nar}, \slkd increases the confidence of source-target attention on both \maskt and \levt. The increase is larger for \maskt than for \levt. For \levt, \slkd increases the attention confidence the most at early decoding iterations. At later iterations, as the model becomes more confident about which source tokens to attend to given the target tokens generated at previous iterations, the impact of \slkd becomes smaller.

Next, we separate the impact of lexical diversity and word reordering (Figure~\ref{fig:attention_complexity}). Reducing both types of complexity leads to more concentrated source-target attention at early iterations. By contrast, models trained on more lexically and syntactically diverse data have more distributed source-target attention at iterations, and the attention becomes more concentrated at later iterations as more target tokens have been generated.

Overall, these results suggest that reducing lexical diversity and degree of word reordering both help \nar find the source-target alignment and thus reduce the error rate at the early decoding stage.

\section{Reduced Lexical Diversity in \slkd Improves Model Confidence}
\label{sec:confidence}

\looseness=-1
\citet{OttAGR2018} show that the intrinsic uncertainty of translation \---\ due to the existence of multiple semantically equivalent translations for the same source \---\ is a source of uncertainty in the \ar models' output distribution.
We hypothesize that these effects might be amplified with \nar models, yet little is known about the confidence and calibration of \nar models. We measure the impact of \slkd on model uncertainty using the average token probability of the models' translation outputs, and the inference Expected Calibration Error~(\ece)~\citep{WangTSL2020} that measures how the model's confidence on a prediction matches to the correctness of the prediction.
As shown in Table~\ref{tab:infece}, both \maskt and \levt become more confident when trained with \slkd. However, \slkd causes \maskt to be overconfident and hurts its calibration by~$+11\%$ \ece.\footnote{This might be due to decoding where \maskt repeatedly masks out and re-predicts its least confident predictions.} By contrast, \slkd changes \levt from underconfident to slightly overconfident, improving its calibration by $-5\%$ lower \ece.

Next, we isolate the impact of lexical diversity and degree of word reordering on model uncertainty.\footnote{We only measure their isolated impact on model uncertainty, not \ece, because we could not isolate lexical diversity from degree of word reordering while controlling faithfulness, which impacts \ece through accuracy.} We measure the average token probability of \maskt and \levt trained on data with varying lexical diversity but close \frs scores~(Figure~\ref{fig:confidence_lex}), and vice versa~(Figure~\ref{fig:confidence_frs}). Decreasing lexical diversity by~$-0.02$ significantly reduces model uncertainty by~$2.1$--$4.6\%$, whereas the impact of word reordering degree is small: increasing \frs by~$+0.08$ only increases the average uncertainty by~$0.8$--$1.5\%$. By contrast, \slkd boosts \frs by~$+0.09$ over the real data. This suggests that reduced lexical diversity is the main reason why \slkd increases model confidence in lexical choice, which raises concerns since \citet{DingWLWTT2021} showed that lexical choice errors are also propagated from \ar to \nar models through \slkd.

%% file: appendix.tex
\appendix

\section{Data Preprocessing Details}
Following \citet{GuWZ2019}, we preprocess the WMT14 En-De and De-En datasets~\citep{BojarWMT2014} via normalization, tokenization, true-casing, and joint BPE~\citep{SennrichHB16bpe} with~$37K$ operations.\footnote{Data can be downloaded from \url{http://dl.fbaipublicfiles.com/nat/original_dataset.zip}} The training data contain~$3.9$M sentence pairs, and the validation and test sets contain~$3,000$ and~$3,003$ sentence pairs, respectively.

\section{Model and Training Details}
\looseness=-1
All \ar and \nar models adopt the \emph{base} Transformer architecture~\citep{Vaswani2017} with~$d_{\text{model}}=512$,~$d_{\text{hidden}}=2048$,~$n_{\text{heads}}=8$,~$n_{\text{layers}}=6$, and~$p_{\text{dropout}} = 0.3$. We tie the source and target embeddings with the output layer weights~\citep{PressW17,NguyenC18}.
We use label smoothing of~$0.1$. We train the models using Adam ~\citep{KingmaB15} with initial learning rate of~$0.0005$ and a batch size of~$64,800$ tokens for maximum~$300,000$ steps. We select the best checkpoint based on validation perplexity. The total number of parameters is~$65$M for the \ar model,~$66$M for \maskt, and~$91$M for \levt. Training takes around~$230$ hours for each \nar model and~$110$ hours for each \ar model on~$4$ Tesla P40 GPUs.

\section{Detailed Experimental Results}
Table~\ref{tab:full_results} shows the scores of corpus-level metrics, test \bleu and validation perplexity of \maskt and \levt trained on various distilled versions of WMT14 De-En training data generated through diverse reference generation~(Section~\ref{sec:method}).

\begin{table*}[ht]
\centering
\scalebox{0.9}{
\begin{tabular}{lcccccrr}
\toprule
& \multicolumn{3}{c}{Data Property} & \multicolumn{2}{c}{test \bleu} & \multicolumn{2}{c}{Valid Perplexity} \\
& \frs & LexDiv & Faith & \maskt & \levt & \maskt & \levt \\
\midrule
Real Data & 0.46 & 0.36 & 0.0 & 28.0 & 27.6 & 35.39 & 62.49 \\
Distilled Data & 0.55 & 0.18 & 7.9 & 29.6 & 30.6 & 8.84 & 11.12 \\
\midrule
Selection: \bleu & 0.54 & 0.19 & 7.4 & 29.4 & 30.1 & 9.74 & 12.57 \\
Selection: \bleu + \nmt score ($\lambda = 0.8$) & 0.54 & 0.18 & 7.4 & 29.2 & 30.1 & 9.45 & 11.94 \\
Selection: \bleu + \nmt score ($\lambda = 0.5$) & 0.55 & 0.17 & 7.6 & 29.5 & 30.6 & 8.97 & 11.59 \\
Selection: \bleu + \nmt score ($\lambda = 0.2$) & 0.55 & 0.17 & 7.8 & 29.2 & 30.4 & 8.77 & 10.94 \\
Selection: \bleu + word-align score ($\lambda = 0.8$) & 0.55 & 0.18 & 7.4 & 29.6 & 30.3 & 9.63 & 12.23  \\
Selection: \bleu + word-align score ($\lambda = 0.5$) & 0.57 & 0.18 & 7.6 & 29.2 & 30.1 & 9.27 & 11.48 \\
Selection: \bleu + word-align score ($\lambda = 0.2$) & 0.58 & 0.18 & 7.9 & 28.7 & 30.0 & 8.69 & 11.24 \\
Selection: \bleu + \frs ($\lambda = 0.8$) & 0.56 & 0.19 & 7.4 & 29.1 & 30.3 & 9.68 & 12.10 \\
Selection: \bleu + \frs ($\lambda = 0.5$) & 0.61 & 0.19 & 7.6 & 28.8 & 29.6 & 9.53 & 12.25 \\
Selection: \bleu + \frs ($\lambda = 0.2$) & 0.64 & 0.19 & 7.8 & 28.5 & 29.7 & 8.81 & 11.71 \\
\hline
\end{tabular}}
\caption{\frs, lexical diversity~(\textit{LexDiv}), and faithfulness~(\textit{Faith}) scores of various distilled versions of WMT14 De-En training data, test \bleu scores and validation perplexity of \maskt and \levt trained on each data version.}
\label{tab:full_results}
\end{table*}

\section{Reference Generation Examples}
We show that the $k$-best list generated by the \ar model using beam search is both lexically and syntactically diverse through a random example selected from the training set~(Table~\ref{tab:example}).

\begin{table*}[!ht]
\centering
\begin{tabular}{ll}
\toprule
source & Ich hoffe , daß dort in Ihrem Sinne entschieden wird. \\
original reference & It will , I hope , be examined in a positive light. \\
\midrule
translation 1 & I hope that it will be decided along your lines. \\
translation 2 & I hope that a decision will be taken along your lines. \\
translation 3 & I hope that the decision will be taken along your lines. \\
translation 4 & I hope that it will be decided in your interest. \\
translation 5 & I hope that there will be a decision along your lines. \\
translation 6 & I hope that decision will be taken along your lines. \\
translation 7 &  I hope that the decision will be taken in your interest. \\
translation 8 & I hope that a decision will be taken in your interest. \\
translation 9 &  I hope that a decision will be made along your lines. \\
translation 10 & I hope that this will be decided along your lines. \\
translation 11 & I hope that a decision will be taken to that effect. \\
translation 12 & I hope there will be a decision along your lines. \\
translation 13 & I hope that a decision will be taken on your behalf. \\
translation 14 & I hope that a decision will be taken in that regard. \\
translation 15 & I hope that decision will be taken in your interest. \\
translation 16 & I hope that a decision will be taken in that direction. \\
translation 17 & I hope that a decision will be taken in that respect. \\
translation 18 & I hope it will be decided along your lines. \\
translation 19 & I hope that you will take a decision there. \\
translation 20 & I hope that you will take a decision in that regard. \\
translation 21 & I hope that this decision will be taken in your interest. \\
translation 22 & I hope that it will decide along your lines. \\
translation 23 & I hope that it will be decided in your interests. \\
translation 24 & I hope that the decision will be taken in your interests. \\
translation 25 & I hope that the decision will be taken in that direction. \\
translation 26 & I hope that a decision will be taken in your interests. \\
translation 27 & I hope that a decision will be taken to that end. \\
translation 28 & I hope that the decision will be taken in that regard. \\
translation 29 & I hope that a decision will be made in your interest. \\
translation 30 & I hope it will be decided in your interest. \\
translation 31 & I hope that you will take a decision on this. \\
translation 32 & I hope that it will be decided accordingly. \\
\bottomrule
\end{tabular} 
\caption{An example of the $k$-best list generated by the \ar model using beam search with a beam size of~$k = 32$.}
\label{tab:example}
\end{table*}